\documentclass{article} 
\usepackage{iclr2016_conference,times}
\usepackage{hyperref}
\usepackage{url}
\usepackage{graphicx,amsmath}

\iclrfinalcopy

\title{Learning Representations from EEG with Deep Recurrent-Convolutional Neural Networks
}

\author{Pouya Bashivan   \\
Electrical and Computer Engineering Department\\
University of Memphis\\
Memphis, TN , USA \\
\texttt{\{pbshivan\}@memphis.edu} \\
\And
Irina Rish \\
IBM T.J. Watson Research Center \\
Yorktown Heights, NY, USA \\
\texttt{\{rish\}@us.ibm.com} \\
\And
Mohammed Yeasin  \\
Electrical and Computer Engineering Department\\
University of Memphis\\
Memphis, TN , USA \\
\texttt{\{myeasin\}@memphis.edu} \\
\And
Noel Codella \\
IBM T.J. Watson Research Center \\
Yorktown Heights, NY, USA \\
\texttt{\{nccodell\}@us.ibm.com} \\
}

\begin{document}

\maketitle

\begin{abstract}
One of the challenges in modeling cognitive events from electroencephalogram (EEG) data is finding representations that are invariant to inter- and intra-subject differences, as well as to inherent noise associated with EEG data collection. Herein, we propose a novel approach for learning such representations from multi-channel EEG time-series, and demonstrate its advantages in the context of mental load classification task. First, we transform EEG activities into a sequence of topology-preserving multi-spectral images, as opposed to standard EEG analysis techniques that ignore such spatial information. Next, we train a deep recurrent-convolutional network inspired by state-of-the-art video classification techniques to learn robust representations from the sequence of images. The proposed approach is designed to preserve the spatial, spectral, and temporal structure of EEG which leads to finding features that are less sensitive to variations and distortions within each dimension. Empirical evaluation on the cognitive load classification task demonstrated significant improvements in classification accuracy over  current state-of-the-art approaches in this field.
\end{abstract}

\section{Introduction}
Deep neural networks have recently achieved great success in recognition tasks within a wide range of applications including images, videos, speech, and text \citep{1-Krizhevsky2012,2-Graves2013,3-Karpathy2014,4-Zhang2015,5-Hermann2015}. Convolutional neural networks (ConvNets) lie at the core of best current architectures working with images and video data, primarily due to their ability to extract representations that are robust to partial translation and deformation of input patterns \citep{6-LeCun1998}. On the other hand, recurrent neural networks have delivered state-of-the-art performance in many  applications involving dynamics in temporal sequences, such as, for example, handwriting and speech recognition \citep{2-Graves2013,7-Graves2008}. In addition, combination of these two network types have recently been used for video classification \citep{8-Ng2015}.

Despite numerous successful applications of deep neural networks to large-scale image, video and text data, they remain relatively unexplored in neuroimaging domain.  Perhaps one of the main reasons here is  that the number of samples in most neuroimaging datasets is limited,  thus making such data less adequate  for training large-scale networks with millions of parameters. As it is often demonstrated,  the advantages of deep neural networks over traditional machine-learning techniques  become more apparent when the dataset size becomes  very large.
Nevertheless, deep belief network and ConvNets have been used to learn representations from functional Magnetic Resonance Imaging (fMRI) and Electroencephalogram (EEG) in some previous work with moderate dataset sizes \citep{9-Plis2014,10-Mirowski2009}. \citet{9-Plis2014} showed that adding several Restricted Boltzman Machine layers to a deep belief network and using supervised pretraining results in networks that can learn increasingly complex representations of the data and achieve considerable accuracy increase  as compared to other classifiers. 
In other works, convolutional and recurrent neural networks have been used to extract representations from EEG time series \citep{10-Mirowski2009, 11-Cecotti2011,12-GULER2005}. These studies demonstrated potential benefits of adopting (down-scaled) deep neural networks in neuroimaging,  even in the absence of  extremely large, million-sample datasets, such as those available for images, video, and text modalities. However, none of these studies attempted to jointly preserve the structure of EEG data within space, time, and frequency.

Herein, we explore the capabilities of deep neural nets for  modeling cognitive events from EEG data. EEG is a widely used noninvasive neuroimaging modality which operates by measuring changes in electrical voltage on the scalp induced by cortical activity. Using the  classical blind-source separation analogy,  EEG data can be thought of as a multi-channel ''speech'' signal obtained from  several  ''microphones'' (associated with EEG electrodes) that record signals from multiple  ''speakers'' (that correspond to activity in  cortical regions). State-of-the-art mental state recognition using EEG consists of manual feature selection from continuous time series and applying  supervised learning algorithms to learn the discriminative manifold between the states \citep{13-Lotte2007,14-Subasi2010}. A key challenge in correctly recognizing  mental states from observed brain activity is constructing  a model that is {\em robust} to translation and deformation of signal in space, frequency, and time, due to inter- and intra-subject differences, as well as signal acquisition protocols. Much of the variations originate from slight individual differences in cortical mapping and/or functioning, giving rise to observed differences in spatial, spectral, and temporal patterns.
Moreover, EEG caps which are used to place the electrodes on top of predetermined cortical regions  can be another source of spatial variations in observed responses due to imperfect fitting of the cap on heads of different sizes and shapes. An example illustrating  potentially high inter- and intra-subject variability in EEG data is given in Appendix.

We propose a novel approach to learning  representations from EEG data that relies on deep learning and appears to be more robust to inter- and intra-subject differences, as well as to measurement-related noise.
Our approach is fundamentally different from the previous attempts to  learn high-level representations from EEG using deep neural networks.
Specifically, rather than representing low-level EEG features as a vector, we transform the data into a multi-dimensional tensor which retains the structure of the data throughout the learning process.
In other words, we obtain a  sequence of topology-preserving multi-spectral images, as opposed to standard EEG analysis techniques that ignore such spatial information. Once such  EEG ''movie'' is obtained, we train  deep recurrent-convolutional neural network  architectures, inspired by state-of-the-art video classification \citep{8-Ng2015}, to learn robust representations from the sequence of images, or frames.
More specifically, we use ConvNets to extract spatial and spectral invariant representations from each frame  data, and adopt LSTM network to extract temporal patterns in the frame sequence. Overall, the proposed approach is  designed to preserve the spatial, spectral, and temporal structure of EEG data, and  to extract features that are more robust to variations and distortions within each dimension. Empirical evaluation on the cognitive load classification task demonstrated significant improvements  over current state-of-the-art approaches in this field, reducing  the classification error from  15.3\% (state-of-art on this application)  to 8.9\%.

\section{Our Approach}
\subsection{Making images from EEG time-series}
Electroencephalogram includes multiple time series corresponding to measurements across different spatial locations over the cortex. Similar to speech signals, the most salient features reside in frequency domain, usually studied using spectrogram of the signal. However, as already noted,   EEG signal  has an  additional spatial dimension.
Fast Fourier Transform (FFT) is performed on the time series for each trial to estimate the power spectrum of the signal. Oscillatory cortical activity related to memory operations primarily exists in three frequency bands of theta (4-7Hz), alpha (8-13Hz), and beta (13-30Hz) \citep{15-Bashivan2014,16-Jensen2002}. Sum of squared absolute values within each of the three frequency bands was computed and used as separate measurement for each electrode.

Aggregating spectral measurements for all electrodes to form a feature vector is the standard approach in EEG data analysis. However, this approach clearly ignores the inherent structure of the data in space, frequency, and time. Instead, we propose to   {\em transform the measurements into a 2-D image} to preserve the spatial structure and {\em use multiple color channels to represent the spectral dimension}. Finally, we {\em use the sequence of images derived from consecutive time windows} to account for temporal evolutions in brain activity.

\begin{figure}[tbh]
\begin{center}
\includegraphics[width=5in]{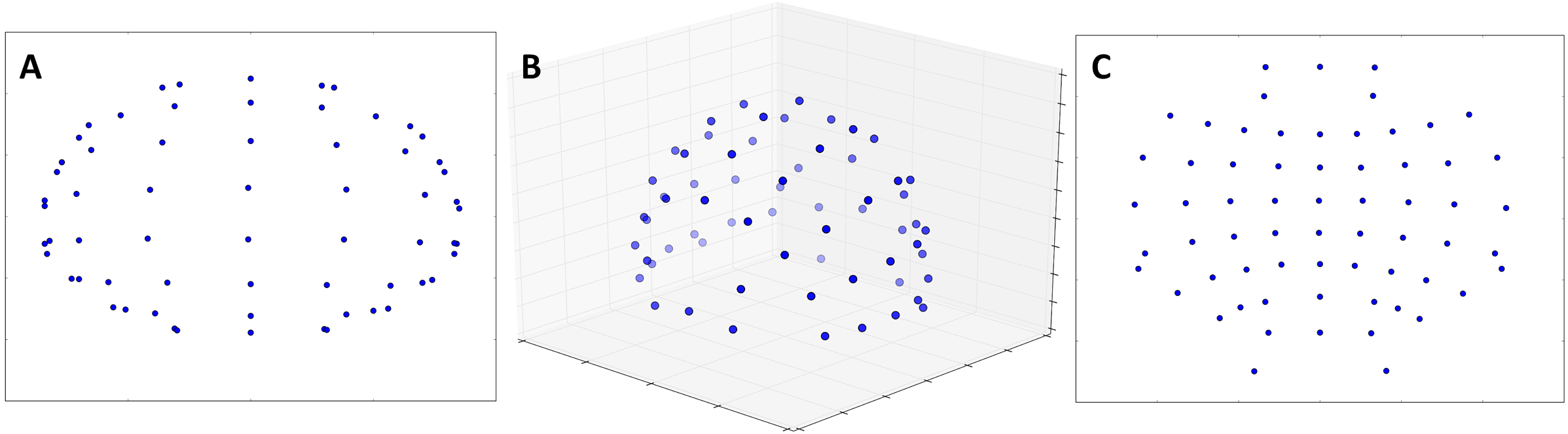}
\end{center}
\caption{Topology-preserving and non-topology-preserving projections of electrode locations. A) 2-D projection of electrode locations using non-topology-preserving simple orthographic projection. B) Location of electrodes in the original 3-D space. C) 2-D projection of electrode locations using topology-preserving azimuthal equidistant projection.}
\label{fig1}
\end{figure}

\begin{figure}[tbh]
\begin{center}
\includegraphics[width=5in]{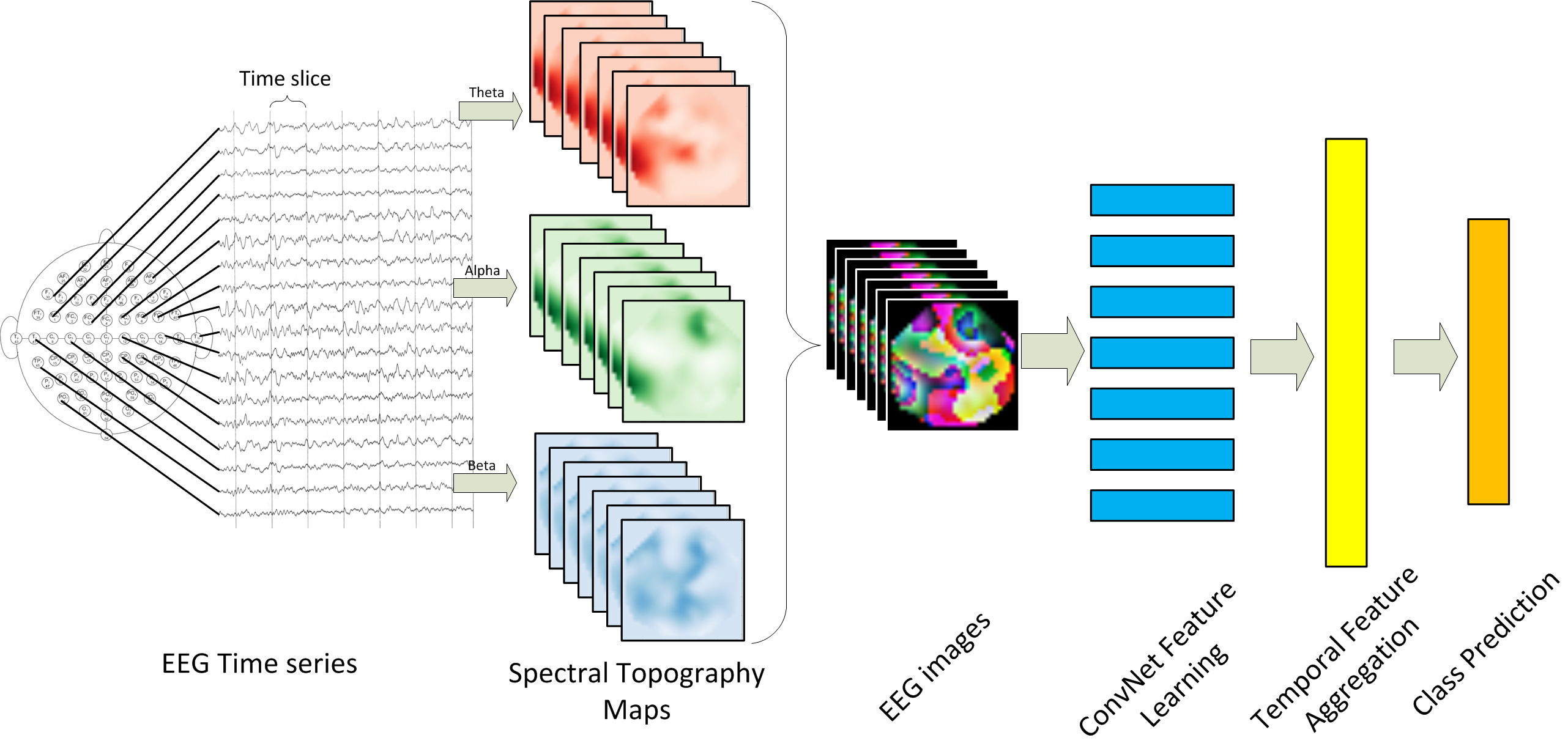}
\end{center}
\caption{Overview of our approach: (1) EEG time series from multiple locations are acquired; (2) spectral power within three prominent frequency bands is extracted for each location and used to form topographical maps for each time frame (image); (3) sequence of topographical maps are combined to form a sequence of 3-channel images which are fed into a recurrent-convolutional network for representation learning and classification.}
\label{fig2}
\end{figure}

The EEG electrodes are distributed over the scalp in a three-dimensional space. In order to transform the spatially distributed activity maps as 2-D images, we need to first project the location of electrodes from a 3-dimensional space onto a 2-D surface. However, such transformation should also preserve the relative distance between neighboring electrodes. For this purpose, we used the Azimuthal Equidistant Projection (AEP) also known as Polar Projection, borrowed from mapping applications \citep{17-Snyder1987}.
The azimuthal projections are formed onto a plane which is usually tangent to the globe at either pole, the Equator, or any intermediate point. In azimuthal equidistant projection, distances from the center of projection to any other point are preserved. Similarly, in our case the shape of the cap worn on a human's head can be approximated by a sphere and the same method could be used to compute the projection of electrode locations on a 2D surface that is tangent to the top point of the head. A drawback of this method is that the distances between the points on the map are only preserved with respect to a single point (the center point) and therefore the relative distances between all pairs of electrodes will not be exactly preserved. Applying AEP to 3-D electrode locations, we obtain 2-D projected locations of electrodes (Figure \ref{fig1}). Width and height of the image represent the spatial distribution of activities over the cortex. We apply Clough-Tocher scheme \citep{alfeld1984trivariate}
for interpolating the scattered power measurements over the scalp and for estimating the values in-between the electrodes over a 32 $\times$ 32 mesh. This procedure is repeated for each frequency band of interest, resulting in three topographical activity maps corresponding to each frequency band. The three spatial maps are then  merged together to form an image with three (color) channels. This three-channel image is given as an  input to a deep convolutional network, as discussed in the following section. Figure \ref{fig2} illustrates an overview of our multi-step approach to mental state classification from EEG data, where the novelty resides in transforming raw EEG into   sequence of images, or frames (EEG ''movie''),
combined with recurrent-convolutional network architecture  applied on top of such transformed EEG data. Note that our approach is general enough to be used in any EEG-based classification task, and a specific problem of mental load classification presented  later only serves as an example demonstrating potential advantages of the proposed approach.

\subsection{ Architecture}
We adopted a recurrent-convolutional neural network to deal with the inherent structure of EEG data. ConvNets were used to deal with variations in space and frequency domains due to their ability to learn good two-dimensional representation of the data. Wherever needed, the extracted representations were  fed into another layer to account for temporal variations in the data. We evaluated various types of layers used for extracting temporal patterns, including convolutional and recurrent layers. Essentially, we evaluated the following two primary approaches to the cognitive state classification problem.  1) {\em Single-frame approach}: a single image was constructed from spectral measurements over the complete trial duration. The constructed image was then used as input to the ConvNet. 2) {\em Multi-frame approach}: We divided each trial into 0.5 second windows and constructed an image over each time window, delivering 7 frames per trial (see section \ref{sec-exp}). The sequence of images was then used as input data to the recurrent-convolutional network. We used Lasagne\footnote{\url{https://github.com/Lasagne/Lasagne}} to implement different architectures discussed in this paper. The code necessary for generating EEG images and building and training the networks discussed in this paper is available online\footnote{\url{https://github.com/pbashivan/EEGLearn}}.

\subsubsection{ConvNet Architecture}
We adopted an architecture mimicking the VGG network used in Imagenet classification challenge \citep{18-Simonyan2015}. This network enjoys a highly scalable architecture which uses stacked convolutional layers with small receptive fields. All convolutional layers use small receptive fields of size 3 $\times$ 3 and stride of 1 pixel with ReLU activation function. The convolution layer inputs are padded with 1 pixel to preserve the spatial resolution after convolution. Multiple convolution layers are stacked together which are followed by maxpool layer. Max-pooling is performed over a 2 $\times$ 2 window with stride of 2 pixels. Number of kernels within each convolution layer increases by a factor of two for layers located in deeper stacks. Stacking of multiple convolution layers leads to effective receptive field of higher dimensions while requiring much less parameters \citep{18-Simonyan2015}.

\subsubsection{Single-Frame Approach}
For this approach the single EEG image was generated by applying FFT on the whole trial duration (3.5 seconds). Purpose of this approach was to find the optimized ConvNet configuration. We first studied a simplified version of the problem by computing the average activity over the complete duration of trial. For this, we computed all power features over the whole duration of trial. Following this procedure, EEG recording for each trial was reduced to a single multi-channel image. We evaluated ConvNet configurations of various depths, as described in Table \ref{table1}.
The convolutional layer parameters here are denoted as conv$<$receptive field size$>$-$<$number of kernels$>$.
Essentially, configuration A involves
only two convolutional layers (Conv3-32) stacked together, followed by maxpool layer; configuration B adds on top of architecture A two more convolutional layers (Conv3-64), followed by another maxpool; then configuration
C adds one more convolutional layer (Conv3-128) followed by maxpool;  configuration D differs from C by  using 4 rather than 2 Conv3-32 convolutional layers at the beginning. Finally,
a fully-connected layer with 512 nodes (FC-512) is added on top of all these architectures, followed by  softmax as the last layer.

\begin{table}[ht]
\caption{Evaluated ConvNet configurations for single-frame approach. The convolutional layer parameters are denoted as conv$<$receptive field size$>$-$<$number of kernels$>$.}
\label{ConvNet-archs}
\begin{center}
\begin{tabular}{c c c c}
\multicolumn{4}{c}{\bf ConvNet Configurations}
\\ \hline \\
A   &B  &C  &D
\\ \hline
\multicolumn{4}{c}{input (32 $\times$ 32 3-channel image)}
\\ \hline
            &           &           &Conv3-32 \\
Conv3-32    &Conv3-32   &Conv3-32   &Conv3-32 \\
Conv3-32    &Conv3-32   &Conv3-32   &Conv3-32 \\
            &           &           &Conv3-32
\\ \hline
\multicolumn{4}{c}{maxpool}
\\ \hline
            &Conv3-64   &Conv3-64   &Conv3-64 \\
            &Conv3-64   &Conv3-64   &Conv3-64
\\ \hline
\multicolumn{1}{c|}{} &\multicolumn{3}{c}{maxpool}
\\ \hline
            &           &Conv3-128  &Conv3-128
\\ \hline
\multicolumn{2}{c|}{} &\multicolumn{2}{c}{maxpool}
\\ \hline \\
\multicolumn{4}{c}{FC-512}
\\ \hline \\
\multicolumn{4}{c}{softmax}
\\ \hline \\
\end{tabular}
\end{center}
\label{table1}
\end{table}

\subsubsection{Multi-Frame Approach}
\begin{figure}[tbh]
\begin{center}
\includegraphics[width=3.6in]{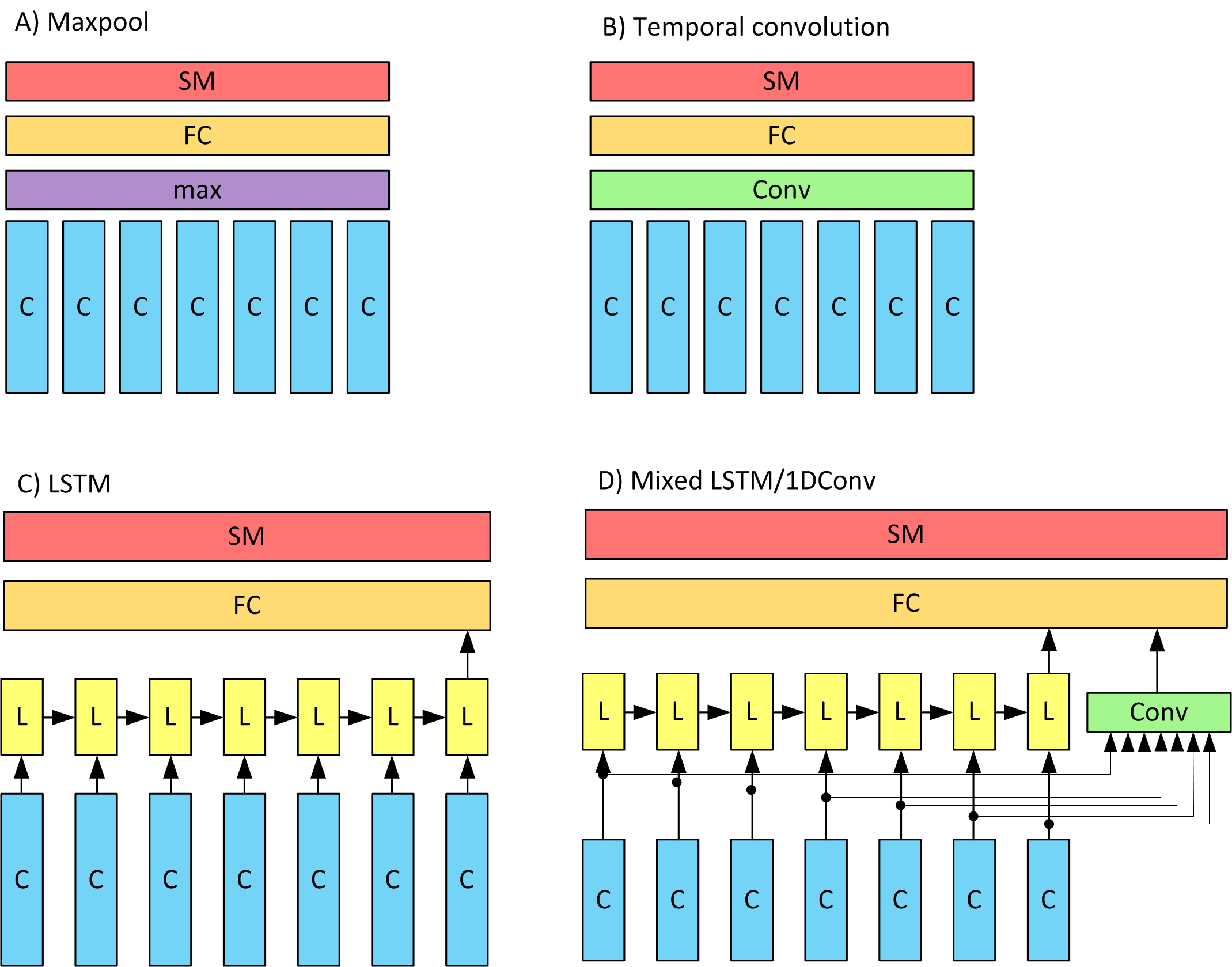}
\end{center}
\caption{Different multi-frame architectures; the notation here is as follow. C: 7-layer ConvNet; max: maxpool layer across time frame features; FC: fully-connected layer; SM: softmax layer; Conv: 1-D convolution layer across time frames; L: LSTM layer.}
\label{fig3}
\end{figure}

We adopted the best performing ConvNet architecture from single frame approach for each frame. In order to reduce the number of parameters in the network, all ConvNets share parameters across frames. Outputs of all ConvNets are reshaped as sequential frames and used to investigate temporal sequence in maps. We evaluated three approaches to extracting temporal information from sequence of activity maps, inspired by a set of deep learning techniques for video classification presented in \citep{8-Ng2015}; see Figure \ref{fig3}: 1) Max-pooling over time; 2) Temporal convolution; 3) LSTM. Finally, the outputs from the last layer are fed to a fully connected layer with 512 hidden units followed by a four-way softmax layer. We kept the number of neurons in the fully connected layer relatively low to control the total number of parameters in the network. 50\% dropout was used on the last two fully connected layers.

\noindent{\bf Max-pooling}: This model performs max-pooling over ConvNet outputs across time frames. While representations found from this model preserve spatial location, they are nonetheless order invariant.

\noindent{\bf Temporal convolution}: This model applies a 1-D convolution to ConvNet outputs across time frames. We evaluated two models consisting of 16 and 32 kernels of size 3 with stride of 1 frame. Kernels capture distinct temporal patterns across multiple frames.

\noindent{\bf Long Short-Term Memory (LSTM)}: Recurrent neural networks take input in the shape of a sequence $x = (x_1,..., x_T)$ and compute hidden vector sequence $h = (h_1,..., h_T)$ and output vector $y = (y_1,..., y_T)$ by iterating the following equations from $t = 1$ to $T$:
\begin{eqnarray}
h_t = H(W_{xh}x_t + W_{hh}h_{t-1} + b_h) \\
y_t = W_{hy}h_t + b_y,
\label{lstm_eq}
\end{eqnarray}
where the  $W$ terms denote weight matrices, $b$ terms denote bias vectors, and $H$ is the hidden layer function.

Given the dynamic nature of neural responses and, consequently, of EEG data, recurrent neural networks (RNN) appear to be a  reasonable choice for modeling temporal evolution of brain activity. Long Short-Term Memory (LSTM) model \citep{19-Hochreiter1997} is a RNN with improved memory. It uses memory cells with an internal memory and gated inputs/outputs which have shown to be more efficient in capturing long-term dependencies. The hidden layer function for LSTM is computed by the following set of equations:
\begin{eqnarray}
i_t = \sigma(W_{xi}x_t + W_{hi}h_{t-1} + W_{ci}c_{t-1} + b_i) \\
f_t = \sigma(W_{xf}x_t+W_{hf}h_{t-1} + W_{cf}c_{t-1} + b_f) \\
c_t = f_t c_{t-1} + i_t \tanh(W_{xc}x_t + W_{hc}h_{t-1} + b_c) \\
o_t = \sigma(W_{xo}x_t + W_{ho}h_{t-1} + W_{co}c_t + b_o) \\
h_t = o_t \tanh(c_t),
\label{lstm_eq2}
\end{eqnarray}
where $\sigma$ is the logistic sigmoid function, and the components of the LSTM model, referred to as {\em input gate}, {\em forget gate}, {\em output gate} and {\em cell activation vectors} are denoted, respectively, as $i$, $f$, $o$, and $c$ (see  \citep{19-Hochreiter1997} for details).

We experimented with up to two LSTM layers and various number of memory cells in each layer and obtained the best results with one layer consisting of 128 cells. Only the prediction made by LSTM after seeing the complete sequence of frames was propagated up to the fully connected layer.

We adopted LSTM to capture temporal evolution in sequences of ConvNet activations. Since brain activity is a temporally dynamic process, variations between frames may contain additional information about the underlying mental state.

\subsection{Training}
Training is carried out by optimizing the cross-entropy loss function. Weight sharing in ConvNets results in vastly different gradients in different layers and for this reason a smaller learning rate is usually used when applying SGD. We trained the recurrent-convolutional network with Adam algorithm \citep{20-Kingma2015} with a learning factor of $10^{-3}$, and decay rate of first and second moments as 0.9 and 0.999 respectively. Batch size was set to 20. Adam has been shown to achieve competitively fast convergence rates when used for training ConvNets as well as multi-layer neural networks. In addition, VGG architecture requires fewer epochs to converge due to implicit regularization imposed by greater depth and smaller convolution filter sizes. The large number of parameters existing in our network made it susceptible to overfitting. We adopted several measures to address the issue. Dropout \citep{23-hinton2012improving} with a probability of 0.5 was used in all fully connected layers. Additionally, we used early stopping by monitoring model's performance over a randomly selected validation set. Dropout regularization has proved to be an effective method for reducing the overfitting in deep neural networks with millions of parameters \citep{1-Krizhevsky2012} and in neuroimaging applications \citep{9-Plis2014}.

\begin{figure}[tbh]
\begin{center}
\includegraphics[width=3.3in]{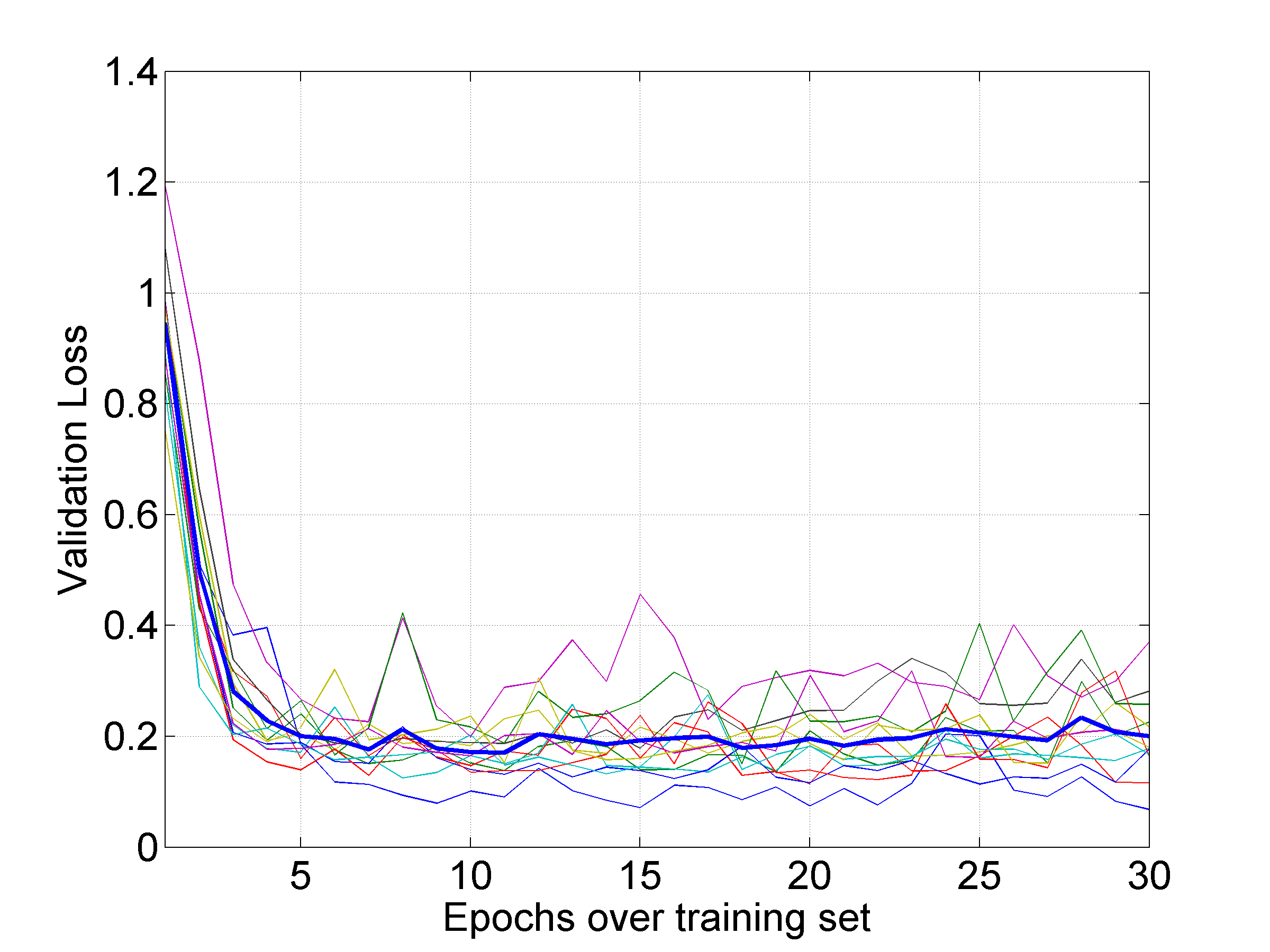}
\end{center}
\caption{Validation loss with training epochs for all cross-validation folds. Thick blue line is the average over all folds.}
\label{fig4}
\end{figure}

Moreover, another commonly used approach for addressing the unbalanced ratio between number of samples and number of model parameters is to artificially expand the dataset using data augmentation. We tried training the network with augmented data generated by randomly adding noise to the images. We did not use image flipping or zooming when augmenting the data due to distinct interpretation of direction and location in EEG images (corresponding to various cortical regions). We experimented with various noise levels added to each image. However, augmenting the dataset did not improve the classification performance and for higher noise values increased the error rates. Figure \ref{fig4} shows the validation loss with number of epochs over the training set. We found that the network parameters converge after about 600 iterations (5 epochs).

\section{Baseline Methods}
 We compared our approach against various classifiers commonly used in the field, including Support-Vector Machines (SVM), Random Forest, sparse Logistic Regression, and Deep Belief Networks (DBN). Here we briefly describe some of the details and parameter settings used in those methods.

\noindent{\bf SVM}: SVM hyperparameters consisting of regularization penalty parameter (C) and inverse of RBF kernel's standard deviation $(\gamma=1/\sigma)$ were selected by grid-search through cross-validation on training set $(C = \{0.01, 0.1, 1, 10, 100\},  \gamma = \{0.1, 0.2,...,1, 2,..., 10\})$.

\noindent {\bf Random Forest}: Random forest is an ensemble method consisting of a group of independent random decision trees. Each tree is grown using a randomly selected subset of features. For each input, outputs of all trees are computed, and the class with majority of votes is selected. The number of estimators for the random forest was varied within the set of $\{5, 10, 20, 50, 100, 500, 1000\}$.

\noindent {\bf Logistic Regression}: $l_1$-regularization was used to introduce sparsity in the logistic regression model. Optimal regularization parameter $C$ was selected via cross-validation on training set, in which the logarithmic range of $[10^{-2}, 10^3]$ was searched.

\noindent {\bf Deep Belief Network}: We used a three-layer Deep Belief Network (DBN). The first layer was a Gaussian-Binary Restricted Boltzman Machine (RBM) and the other two layers were Binary RBMs. The output of the final level was fed into a two-way softmax layer for predicting the class label. Parameters of each layer of DBN were greedily pre-trained to improve learning by shifting the initial random parameter values toward a good local minimum \citep{21-Bengio2007}. We used the following empirically selected numbers of neurons in the three layers that demonstrated good performance: 512, 512, and 128. The last layer was connected to a softmax layer with 4 units. The network was fine-tuned using batch stochastic gradient descent with $l_1$-regularization to reduce the overfitting  during training.

\section{Experiments on an EEG dataset}
\label{sec-exp}
Every individual has a different cognitive processing capacity which causally determines his/her ability in performing mental tasks. While human brain consists of numerous networks responsible for specialized tasks, many of them rely on more basic functional networks like working memory. Working memory is responsible for transient retention of information which is crucial for any manipulation of information in the brain. Its capacity sets bounds on individual's ability in a range of cognitive functions. Increasing cognitive demand (load) beyond individual's capacity leads to overload state causing confusion and diminished learning ability \citep{22-Sweller1998}. For this reason, ability to recognize individual's cognitive load becomes important for many applications including brain-computer interfaces, human-computer interaction, and tutoring services.

\begin{figure}[tbh]
\begin{center}
\includegraphics[width=3.3in]{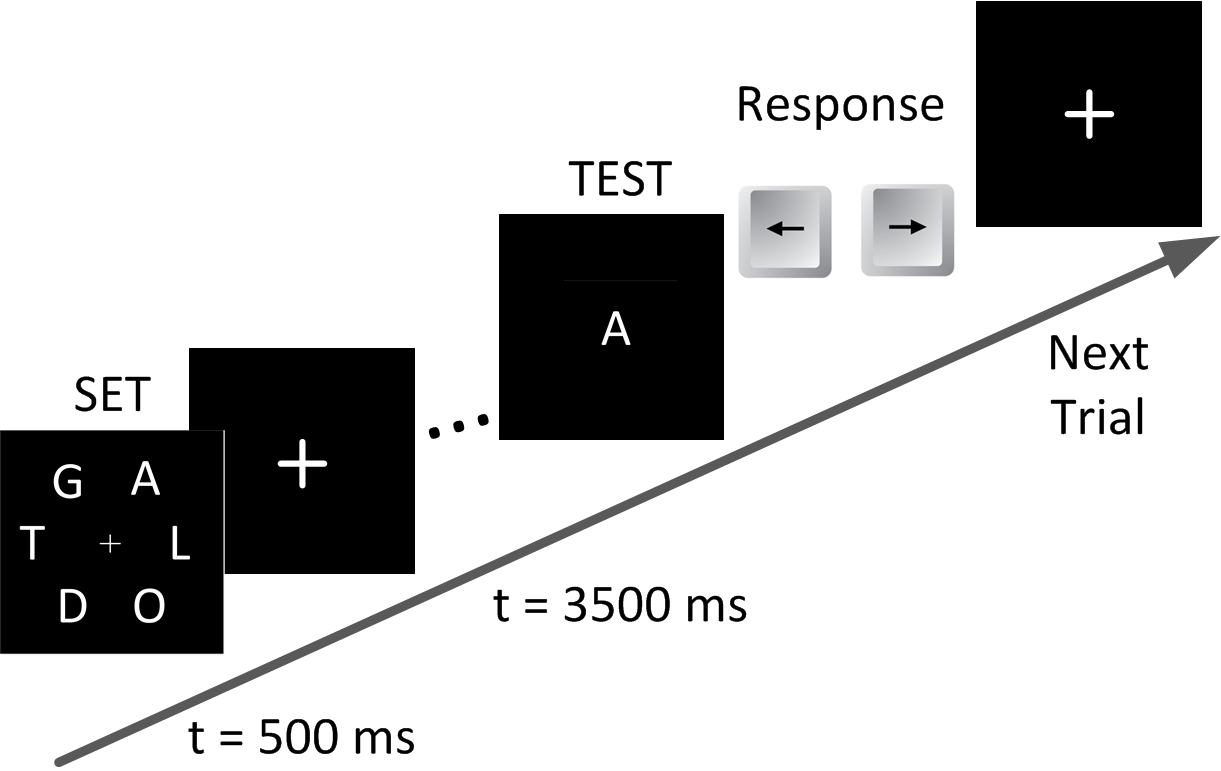}
\end{center}
\caption{Working memory experiment diagram; participants briefly observe an array containing multiple English characters ‘SET’ (500ms) and maintain the information for three seconds. A ‘TEST’ character is then presented and participants respond by press of a button if TEST charter matches one of the characters in the SET.}
\label{fig5}
\end{figure}

Here we used an EEG dataset acquired during a working memory experiment. EEG was recorded as fifteen participants (eight female) performed a standard working memory experiment. Details of procedures for data recording and cleaning are reported in our previous publication \citep{15-Bashivan2014}. In brief, continuous EEG was recorded from 64 electrodes placed over the scalp at standard 10-10 locations with a sampling frequency of 500 Hz. Electrodes are placed at distances of 10\% along the medial-lateral contours. Data for two of the subjects was excluded from the dataset because of excessive noise and artifacts in their recorded data. During the experiment, an array of English characters was shown for 0.5 second (‘SET’) and participants were instructed to memorize the characters. A ‘TEST’ character was shown three seconds later and participants indicated whether the test character was among the first array ('SET') or not by press of a button. Each participant repeated the experiment for 240 times. The number of characters in the ‘SET’ for each trial was randomly chosen to be 2, 4, 6, or 8. The number of characters in the ‘SET’ determines the amount of cognitive load induced on the participant as with increasing number of characters more mental resources are required to retain the information. Throughout the paper, we identify each of the conditions containing {2, 4, 6, 8} characters with loads 1-4 respectively. Recorded brain activity during the period which individuals retained the information in their memory (3.5 seconds) was used to recognize the amount of mental workload. Figure \ref{fig5} demonstrates the time course of the working memory experiment. The classification task is to recognize the load level corresponding to set size (number of characters presented to the subject) from EEG recordings. Four distinct classes corresponding to load 1-4 are defined and the 2670 samples collected from 13 subjects are assigned to these four categories.

Continuous EEG was sliced offline to equal lengths of 3.5 seconds corresponding to each trial. A total of 3120 trials were recorded. Only data corresponding to correctly responded trials were included in the data set which reduced the data set size to 2670 trials. For evaluating the performance of each classifier we followed the leave-subject-out cross validation approach. In each of the 13 folds, all trials belonging to one of the subjects were used as the test set. A number of samples equal to the test set were then randomly extracted from rest of data for validation set and the remaining samples were used as training set.

\section{Results}
We examined the EEG dataset from two approaches. In the first approach (single-frame) we extracted the power features by applying FFT on the complete duration of each trial leading to single 3-channel image corresponding to each trial. The second approach included dividing each trial to multiple time windows and extracting power features for each window separately leading to conservation of temporal information rather than averaging them out into single slice of activity map.

\subsection{Single-frame classification}
We first present our results on classification using a single frame derived by extracting features over the complete trial duration and applying ConvNets. The purpose of this part was to empirically seek the best performing ConvNet architecture working on images generated from complete EEG time series. We evaluated various configurations with different number of convolution and maxpool layers. We followed the VGG architecture for selection of number of filters in each layer and grouping convolution layers with small receptive fields.

Table \ref{table1} presented earlier summarized the architectures we considered.   Table \ref{table2} shows the number of parameters used by each type of architecture, and the corresponding error achieved on the test set. We found ConvNet based architectures to be superior to our baseline methods. We can see that increasing the number of layers to seven slightly improved the achievable error rates on the test set. The best result was obtained with architecture ‘D’ containing 7 convolution layers which was also marginally better than the baseline methods. While the difference in the error rates between the four configurations was not statistically significant, we chose architecture ‘D’ because of its equal or better error rates on the subset of subjects that were considered hard to classify (up to 12$\%$ decrease in error rates). Most of the network parameters lie in the last two layers (fully connected and softmax) containing approximately 1 million parameters. In VGG style network, the number of filters in each layer is selected in a way that size of the output remains the same after each stack (filter size $\times$ number of kernels).

To quantify the importance of projection type on our results, we additionally generated the images using a simple orthographic projection (onto the z=0 plane) and retrained our network. The differences between topology-preserving and non-topology-preserving projections were mostly evident on the peripheral parts of the projected image (Figure \ref{fig1}). In our experiments we observed slight improvement of classification error in using topology preserving projection over non-equidistant flattening projection ($\sim$0.6$\%$). However, this observation could be dependent on the particular dataset and requires further exploration to conclude. Moreover, using the equidistant projection approach helps with the interpretability of images and feature maps when visualizing the data. Overall, our claim is that mapping EEG data into a 2D image (specially with equidistant projections) leads to considerably better classification of cognitive load levels as compared to standard, non-spatial approaches that treat EEG simply as a collection of time series.

\begin{table}[ht]
\caption{The number of parameters in convolutional layers for evaluated single-frame architectures, and the test errors achieved by each architecture.}
\label{ConvNet-numpars}
\begin{center}
\begin{tabular}{c c c}
\multicolumn{1}{c}{\bf Model}  &\multicolumn{1}{c}{\bf Number of parameters}
&\multicolumn{1}{c}{\bf Test Error (\%)}
\\ \hline \\
\bf RBF SVM                     &-      &14.68      \\
\bf L1- Logistic Regression     &-      &14.55      \\
\bf Random Forest               &-      &14.44      \\
\bf DBN                         &428k   &13.59      \\
\bf ConvNet Arch. A             &10k    &13.05      \\
\bf ConvNet Arch. B             &65.5k  &13.17      \\
\bf ConvNet Arch. C             &139.4k &13.91      \\
\bf ConvNet Arch. D             &158k   &\bf12.39      \\
\end{tabular}
\end{center}
\label{table2}
\end{table}


\subsection{Multi-frame classification}
\label{sec-results_MF}
For the multi-frame classification, we used ConvNet with architecture D from previous step and applied it on each frame. We explored the four different approaches to aggregate temporal features from multiple frames (Figure \ref{fig3}). Using temporal convolution and LSTM significantly improved the classification accuracy (see Table \ref{table3}). For the model with temporal convolution, we found the network consisting of 32 kernels to outperform the one with 16 kernels (11.32$\%$ Vs. 12.86$\%$ error). A closer look at the accuracies derived for each individual, reveals that while both methods are achieving close to perfect classification accuracies for eight of participants, most of the differences originated from differences in accuracy for the remaining five individuals (Table \ref{table4}). This observation motivated us to use a combination of temporal convolution and LSTM structures together in single structure which led to our best results on the dataset.

\begin{table}[b]
\begin{small}
\caption{Classification results for multi-frame approach}
\label{multiframe-results}
\begin{center}
\begin{tabular}{c c c c}
\multicolumn{1}{c}{\bf Architecture}  &\multicolumn{1}{c}{\bf Test Error (\%)}
&\multicolumn{1}{c}{\bf Validation Error (\%)} &\multicolumn{1}{c}{\bf Number of parameters}
\\ \hline \\
\bf RBF SVM                 &15.34  &-      &-          \\
\bf L1-Logistic Regression  &15.32  &-      &-          \\
\bf Random Forest           &12.59  &-      &-          \\
\bf DBN                     &14.96  &8.37   &1.02 mil   \\
\bf ConvNet+Maxpool         &14.80  &8.48   &1.21 mil   \\
\bf ConvNet+1D-Conv         &11.32  &9.28   &441 k      \\
\bf ConvNet+LSTM            &10.54  &6.10   &1.34 mil   \\
\bf ConvNet+LSTM/1D-Conv    &\bf8.89&8.39 &1.62 mil \\
\end{tabular}
\end{center}
\label{table3}
\end{small}
\end{table}

\begin{table}[t]
\begin{scriptsize}
\caption{Classification results for each fold}
\label{results-folds}
\begin{center}
\begin{tabular}{c c c c c c c c c c c c c c}
\multicolumn{1}{c}{\bf Test Subject}  &\multicolumn{1}{c}{\bf S1} &\multicolumn{1}{c}{\bf S2}
&\multicolumn{1}{c}{\bf S3} &\multicolumn{1}{c}{\bf S4} &\multicolumn{1}{c}{\bf S5}
&\multicolumn{1}{c}{\bf S6} &\multicolumn{1}{c}{\bf S7} &\multicolumn{1}{c}{\bf S8}
&\multicolumn{1}{c}{\bf S9} &\multicolumn{1}{c}{\bf S10} &\multicolumn{1}{c}{\bf S11}
&\multicolumn{1}{c}{\bf S12} &\multicolumn{1}{c}{\bf S13}
\\ \hline \\
\bf 1D-Conv     &88.3  &72.5    &\bf93.9 &97.5   &98.3   &98     &98.2   &\bf100 &98.5   &94.5   &88.5   &79.5   &45.9      \\
\bf LSTM        &56.7  &73.5    &92.2    &\bf99  &99.4   &\bf99.5&98.9   &\bf100 &\bf100 &\bf97.7&\bf99  &88     &\bf59.1   \\
\bf Mix         &\bf88.9&\bf76.5&93.3    &\bf99  &\bf100 &98     &\bf100 &98.5   &99     &96.8   &96.5   &\bf91  &46.8      \\
\end{tabular}
\end{center}
\label{table4}
\end{scriptsize}
\end{table}

Comparing the performance of baseline models in multi-frame and single-frame cases, test errors were slightly lower for the single-frame setup in all classifiers except random forest. This difference is mostly because of the negative effect of increased number of features in multi-frame case which happens despite the existence of regularization term in all baseline methods. On the other hand, incorporating temporal dynamics (multiple frames over time) in our model increasingly improves the classification performance which demonstrates the effectiveness of our model in learning from time-dependent changes. Moreover, while our approach does not directly operate on raw EEG time-series, we drastically reduced the amount of required data by manually extracting power features from EEG. In addition, discovering complex temporal relationships such as those related to spectral properties in time-series using neural networks, is still an open question which has not been fully addressed.

ConvNets attain translation invariance through maxpooling which is a downsampling procedure in nature. While this helps with creating invariant (with respect to space and frequency) feature maps in the deeper layers of ConvNet, it might also hurt the performance if the feature map size is reduced to a degree in which the regional activities cannot be distinguished from each other. In a sense, there is a trade-off between the degree of abstraction realized through layers of convolution and maxpooling and the level of detail kept in the feature maps. In addition, ConvNets learn stack of filters which produce nonlinear feature maps maximizing the classification accuracy. When trained on a pool of data containing multiple individuals, the network extracts features that are maximally informative considering the variability in the training set.

We note that performance of ConvNet+Maxpool is lower than ConvNet in single-frame setup. Temporal maxpool selects the highest activation across the frames whereas features extracted in the single-frame approach are similar to average values over multiple frames. Choosing the maximum value over multiple time frames is not necessarily the best practice when dealing with brain activity time series as it will potentially ignore the periods of inactivation in some cortical regions. This effect is still partially observable when computing the average of activities over all the frames. It also partially explains lower classification errors when temporal dynamic models (1D-conv and LSTM) are added to the network.

\subsection{Visualizing the learned representations}
The recurrent-convolutional network in section \ref{sec-results_MF} significantly reduced the classification error rates in comparison to all baseline methods through automatic learning of representations from the sequence of EEG images. Understanding how this model achieves such performance is of equal importance. Viewing the learned kernels as images is a classic approach for understanding the learned representations by the network. However, in our network, due to the small reception field dimension of kernels (3$\times$3), displaying the kernels would not give much intuition about the learned representations.
Instead, we adopted deconvolutional network (deconvnet) \citep{Zeiler2011, Zeiler2014, zeiler2010deconvolutional} to visualize the model's learned filters by back-propagating the feature map onto the input space. Deconvnet iteratively approximates the convnet features in previous layer and collectively projects a particular feature map to the input space. This reveals the structures in the input space that excite that particular feature map. To approximately invert the convolution operation, transpose of filters is used instead. The transposed filter is applied to the rectified map at each stage. Maxpool layers are inverted through a bicubic interpolation operation. We computed the back projections of feature maps derived from the last convolution layer of each stack (corresponding to convolution layers 4, 6, and 7 in architecture 'D') for all training images.

\begin{figure}[tbh]
\begin{center}
\includegraphics[width=3.3in]{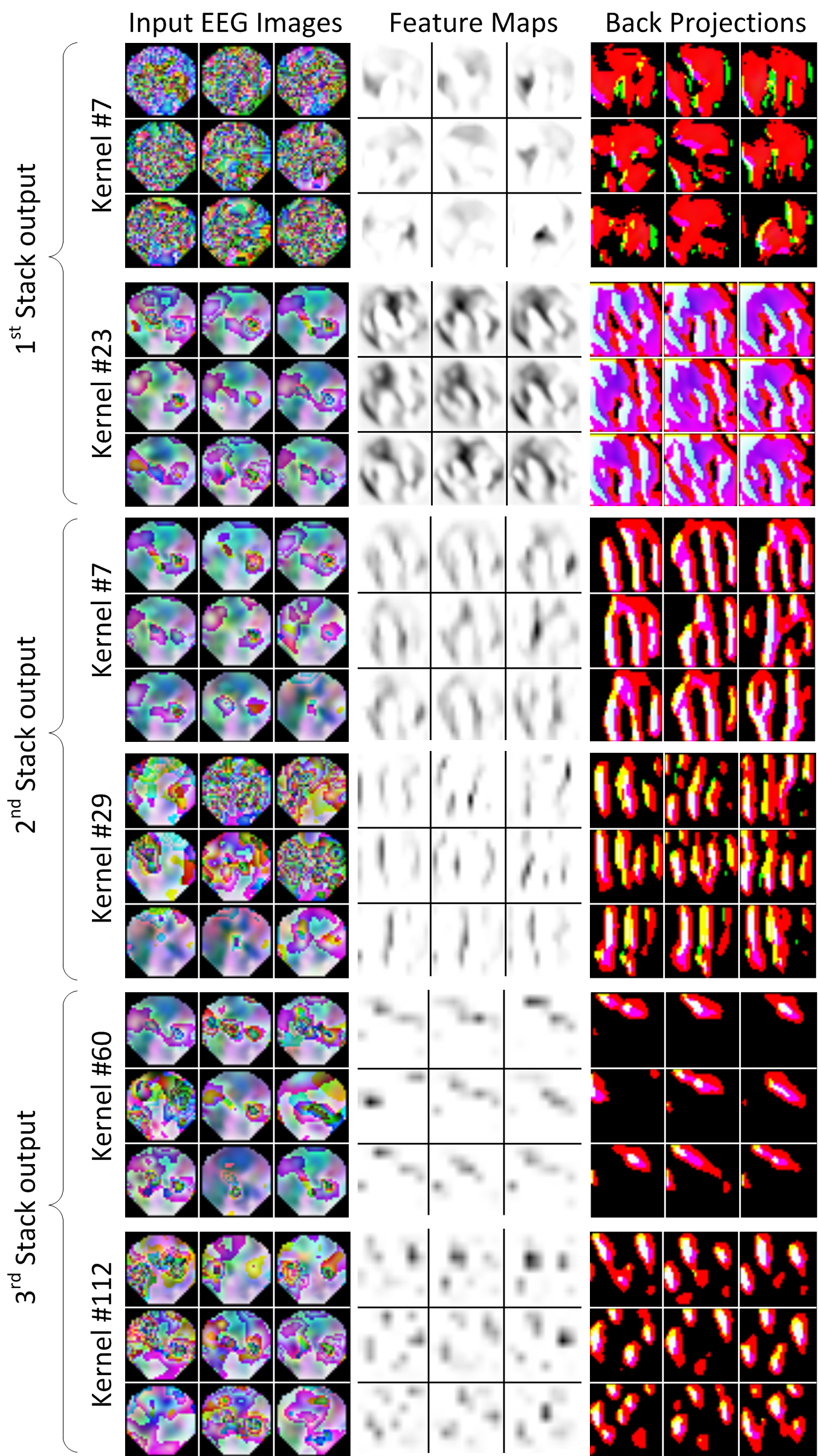}
\end{center}
\caption{Visualization of feature maps and their input activation patterns at various depth levels of convolutional network. The left column (Input EEG images) shows the top 9 images with highest feature activations across the training set. The middle column (Feature Maps) shows the feature map derived in the output of the particular kernel. Right column (Back Projections) shows the back-projected maps derived by applying deconvnet on the feature map displaying structures in the input image that excite that particular feature map.
}
\label{fig6}
\end{figure}

Generally, the lower-level feature maps had a more wide-spread input activation area, while for deeper-layer feature maps the activation areas became sparser. There was also strong frequency selectivity in many of the learned filters. We found some of these features to have noticeable links to well-known electrophysiological markers of cognitive load. Frontal theta and beta activity as well as parietal alpha are most prominent markers of cognitive/memory load in neuroscience literature \citep{Bashivan2015a, Jensen2002, Onton2005, tallon1999sustained}. Figure \ref{fig6} illustrates the back-projected maps for a number of filters with clear neuroscientific interpretations selected from different depths of the network. For each filter we showed the input image, filter output and the back projected activation for 9 images with highest activations (average activation over all feature map pixels) across the training set. Among the  first-layer features we found one feature map capturing wide-spread theta (1\textsuperscript{st} stack output-kernel7) and another frontal beta activity (1\textsuperscript{st} stack output-kernel23). In the second- and third-layer features we observed detectors of frontal theta/beta (2\textsuperscript{nd} stack output-kernel7 and 3\textsuperscript{rd} stack output-kernel60, 112) as well as parietal alpha (2\textsuperscript{nd} stack output-kernel29) with increasing focal specificity of feature maps in deeper layers. The similarity between feature maps derived from different images is noticeable despite considerable dissimilarity in the original input images.

\section{Conclusions}

This work is motivated by the high-level goal of finding robust representations from EEG data, that would be invariant to inter- and intra-subject differences and to inherent noise associated with EEG data collection. We propose a novel methodology  for learning  representations from multi-channel EEG time-series, and demonstrate its advantages in the context of mental load classification task. Our approach is fundamentally different from the previous attempts to  learn high-level representations from EEG using deep neural networks.
Specifically, rather than representing low-level EEG features as a vector, we transform the data into a sequence of topology-preserving multi-spectral images (EEG ''movie''), as opposed to standard EEG analysis techniques that ignore such spatial information.  We then  train deep recurrent-convolutional networks inspired by state-of-the-art video classification to learn robust representations from the sequence of images. The proposed approach demonstrates significant improvements in classification accuracy over the state-of-the-art results.
Since our approach transforms the EEG data into sequence of EEG images, it can be applied on EEG data acquired with different hardware (e.g. with different number of electrodes). The preprocessing step used in our approach transforms the EEG time-series acquired from various sources into comparable EEG frames. In this way, various EEG datasets could be merged together. The only information needed to complete this transform would be the spatial coordinates of electrodes for each setup. As a future direction, it would be possible to use unsupervised pretraining methods with larger (or merged) unlabeled EEG datasets prior to training the network with task-specific data.

\bibliography{iclr2016_conference}

\begin{thebibliography}{31}
\providecommand{\natexlab}[1]{#1}
\providecommand{\url}[1]{\texttt{#1}}
\expandafter\ifx\csname urlstyle\endcsname\relax
  \providecommand{\doi}[1]{doi: #1}\else
  \providecommand{\doi}{doi: \begingroup \urlstyle{rm}\Url}\fi

\bibitem[Alfeld(1984)]{alfeld1984trivariate}
Alfeld, Peter.
\newblock A trivariate clough—tocher scheme for tetrahedral data.
\newblock \emph{Computer Aided Geometric Design}, 1\penalty0 (2):\penalty0
  169--181, 1984.

\bibitem[Bashivan et~al.(2014)Bashivan, Bidelman, and Yeasin]{15-Bashivan2014}
Bashivan, Pouya, Bidelman, Gavin~M., and Yeasin, Mohammed.
\newblock {Spectrotemporal dynamics of the EEG during working memory encoding
  and maintenance predicts individual behavioral capacity}.
\newblock \emph{European Journal of Neuroscience}, 40\penalty0 (12):\penalty0
  3774--3784, 2014.

\bibitem[Bashivan et~al.(2015)Bashivan, Yeasin, and M.]{Bashivan2015a}
Bashivan, Pouya, Yeasin, Mohammed, and M., Bidelman~Gavin.
\newblock Single trial prediction of normal and excessive cognitive load
  through eeg feature fusion.
\newblock \emph{Proceedings of IEEE Signal Processing in Medicine and Biology
  (SPMB) conference}, pp.\  1--5, December 2015.

\bibitem[Bengio et~al.(2007)Bengio, Lamblin, Popovici, and
  Larochelle]{21-Bengio2007}
Bengio, Yoshua, Lamblin, Pascal, Popovici, Dan, and Larochelle, Hugo.
\newblock {Greedy layer-wise training of deep networks}.
\newblock \emph{Advances in neural information processing systems},
  19:\penalty0 153, 2007.
\newblock ISSN 1049-5258.

\bibitem[Cecotti \& Gr{\"{a}}ser(2011)Cecotti and Gr{\"{a}}ser]{11-Cecotti2011}
Cecotti, Hubert and Gr{\"{a}}ser, Axel.
\newblock {Convolutional neural networks for P300 detection with application to
  brain-computer interfaces}.
\newblock \emph{IEEE Transactions on Pattern Analysis and Machine
  Intelligence}, 33\penalty0 (3):\penalty0 433--445, 2011.
\newblock ISSN 01628828.
\newblock \doi{10.1109/TPAMI.2010.125}.

\bibitem[Graves et~al.(2008)Graves, Fern{\'{a}}ndez, and Liwicki]{7-Graves2008}
Graves, Alex, Fern{\'{a}}ndez, S, and Liwicki, Marcus.
\newblock {Unconstrained online handwriting recognition with recurrent neural
  networks}.
\newblock \emph{Advances in Neural Information Processing Systems},
  20:\penalty0 1--8, 2008.

\bibitem[Graves et~al.(2013)Graves, Mohamed, and Hinton]{2-Graves2013}
Graves, Alex, Mohamed, Abdel-Rahman, and Hinton, Geoffrey~E.
\newblock {Speech recognition with deep recurrent neural networks}.
\newblock \emph{Acoustics, Speech and Signal Processing (ICASSP), 2013 IEEE
  International Conference on}, pp.\  6645--6649, 2013.
\newblock ISSN 1520-6149.
\newblock \doi{10.1109/ICASSP.2013.6638947}.

\bibitem[Guler et~al.(2005)Guler, Ubeyli, and Guler]{12-GULER2005}
Guler, N, Ubeyli, E, and Guler, I.
\newblock {Recurrent neural networks employing Lyapunov exponents for EEG
  signals classification}.
\newblock \emph{Expert Systems with Applications}, 29\penalty0 (3):\penalty0
  506--514, 2005.
\newblock ISSN 09574174.
\newblock \doi{10.1016/j.eswa.2005.04.011}.

\bibitem[Hermann et~al.(2015)Hermann, Ko{\v{c}}isk{\'{y}}, Grefenstette,
  Espeholt, Kay, Suleyman, and Blunsom]{5-Hermann2015}
Hermann, Karm~Moritz, Ko{\v{c}}isk{\'{y}}, Tom{\'{a}}{\v{s}}, Grefenstette,
  Edward, Espeholt, Lasse, Kay, Will, Suleyman, Mustafa, and Blunsom, Phil.
\newblock {Teaching Machines to Read and Comprehend}.
\newblock \emph{arXiv}, pp.\  1--13, 2015.

\bibitem[Hinton et~al.(2012)Hinton, Srivastava, Krizhevsky, Sutskever, and
  Salakhutdinov]{23-hinton2012improving}
Hinton, Geoffrey~E, Srivastava, Nitish, Krizhevsky, Alex, Sutskever, Ilya, and
  Salakhutdinov, Ruslan~R.
\newblock Improving neural networks by preventing co-adaptation of feature
  detectors.
\newblock \emph{arXiv preprint arXiv:1207.0580}, 2012.

\bibitem[Hochreiter \& Schmidhuber(1997)Hochreiter and
  Schmidhuber]{19-Hochreiter1997}
Hochreiter, Sepp and Schmidhuber, J{\"{u}}rgen.
\newblock {Long Short-Term Memory}.
\newblock \emph{Neural Computation}, 9\penalty0 (8):\penalty0 1735--1780, 1997.
\newblock ISSN 0899-7667.
\newblock \doi{10.1162/neco.1997.9.8.1735}.

\bibitem[Jensen \& Tesche(2002)Jensen and Tesche]{16-Jensen2002}
Jensen, Ole and Tesche, Claudia~D.
\newblock {Frontal theta activity in humans increases with memory load in a
  working memory task}.
\newblock \emph{Neuroscience}, 15\penalty0 (8):\penalty0 1395--1399, 2002.
\newblock ISSN 0953816X.
\newblock \doi{10.1046/j.1460-9568.2002.01975.x}.

\bibitem[Jensen et~al.(2002)Jensen, Gelfand, Kounios, and Lisman]{Jensen2002}
Jensen, Ole, Gelfand, Jack, Kounios, John, and Lisman, John~E.
\newblock {Oscillations in the alpha band (9-12 Hz) increase with memory load
  during retention in a short-term memory task.}
\newblock \emph{Cerebral cortex (New York, N.Y. : 1991)}, 12\penalty0
  (8):\penalty0 877--82, aug 2002.
\newblock ISSN 1047-3211.

\bibitem[Karpathy \& Toderici(2014)Karpathy and Toderici]{3-Karpathy2014}
Karpathy, Andrej and Toderici, G.
\newblock {Large-scale video classification with convolutional neural
  networks}.
\newblock \emph{Computer Vision and Pattern Recognition (CVPR)}, pp.\
  1725--1732, 2014.
\newblock \doi{10.1109/CVPR.2014.223}.

\bibitem[Kingma \& Ba(2015)Kingma and Ba]{20-Kingma2015}
Kingma, Diederik~P. and Ba, Jimmy~Lei.
\newblock {Adam: a Method for Stochastic Optimization}.
\newblock \emph{International Conference on Learning Representations}, pp.\
  1--13, 2015.

\bibitem[Krizhevsky et~al.(2012)Krizhevsky, Sutskever, and
  Hinton]{1-Krizhevsky2012}
Krizhevsky, Alex, Sutskever, Ilya, and Hinton, Geoffrey~E.
\newblock {Imagenet classification with deep convolutional neural networks}.
\newblock \emph{Advances in neural information processing systems}, pp.\
  1097--1105, 2012.
\newblock ISSN 10495258.

\bibitem[LeCun et~al.(1998)LeCun, Bottou, Bengio, and Haffner]{6-LeCun1998}
LeCun, Yann, Bottou, L{\'{e}}on, Bengio, Yoshua, and Haffner, Patrick.
\newblock {Gradient-based learning applied to document recognition}.
\newblock \emph{Proceedings of the IEEE}, 86\penalty0 (11):\penalty0
  2278--2323, 1998.
\newblock ISSN 00189219.
\newblock \doi{10.1109/5.726791}.

\bibitem[Lotte \& Congedo(2007)Lotte and Congedo]{13-Lotte2007}
Lotte, F and Congedo, M.
\newblock {A Review of Classification Algorithms for EEG-based Brain-Computer
  Interfaces}.
\newblock \emph{Journal of neural engineering}, 4:\penalty0 1--24, 2007.

\bibitem[Mirowski et~al.(2009)Mirowski, Madhavan, LeCun, and
  Kuzniecky]{10-Mirowski2009}
Mirowski, Piotr, Madhavan, Deepak, LeCun, Yann, and Kuzniecky, Ruben.
\newblock {Classification of patterns of EEG synchronization for seizure
  prediction}.
\newblock \emph{Clinical Neurophysiology}, 120\penalty0 (11):\penalty0
  1927--1940, 2009.
\newblock ISSN 13882457.
\newblock \doi{10.1016/j.clinph.2009.09.002}.

\bibitem[Ng et~al.(2015)Ng, Hausknecht, Vijayanarasimhan, Vinyals, Monga, and
  Toderici]{8-Ng2015}
Ng, Jyh, Hausknecht, M, Vijayanarasimhan, S, Vinyals, O, Monga, R, and
  Toderici, G.
\newblock {Beyond Short Snippets: Deep Networks for Video Classification}.
\newblock In \emph{CVPR}, 2015.
\newblock ISBN 9781467369640.

\bibitem[Onton et~al.(2005)Onton, Delorme, and Makeig]{Onton2005}
Onton, Julie, Delorme, Arnaud, and Makeig, Scott.
\newblock {Frontal midline EEG dynamics during working memory.}
\newblock \emph{NeuroImage}, 27\penalty0 (2):\penalty0 341--56, aug 2005.
\newblock ISSN 1053-8119.
\newblock \doi{10.1016/j.neuroimage.2005.04.014}.

\bibitem[Plis et~al.(2014)Plis, Hjelm, Salakhutdinov, Allen, Bockholt, Long,
  Johnson, Paulsen, Turner, and Calhoun]{9-Plis2014}
Plis, Sergey~M., Hjelm, Devon~R., Salakhutdinov, Ruslan, Allen, Elena~a.,
  Bockholt, Henry~J., Long, Jeffrey~D., Johnson, Hans~J., Paulsen, Jane~S.,
  Turner, Jessica~a., and Calhoun, Vince~D.
\newblock {Deep learning for neuroimaging: a validation study}.
\newblock \emph{Frontiers in Neuroscience}, 8\penalty0 (August):\penalty0
  1--11, 2014.
\newblock ISSN 1662-453X.
\newblock \doi{10.3389/fnins.2014.00229}.

\bibitem[Simonyan \& Zisserman(2015)Simonyan and Zisserman]{18-Simonyan2015}
Simonyan, K and Zisserman, A.
\newblock {Very Deep Convolutional Networks for Large-Scale Image Recognition}.
\newblock In \emph{ICLR}, pp.\  1--14, 2015.

\bibitem[Snyder(1987)]{17-Snyder1987}
Snyder, John~Parr.
\newblock \emph{{Map projections--A working manual}}, volume 1395.
\newblock US Government Printing Office, 1987.

\bibitem[Subasi \& {Ismail Gursoy}(2010)Subasi and {Ismail
  Gursoy}]{14-Subasi2010}
Subasi, Abdulhamit and {Ismail Gursoy}, M.
\newblock {EEG signal classification using PCA, ICA, LDA and support vector
  machines}.
\newblock \emph{Expert Systems with Applications}, 37\penalty0 (12):\penalty0
  8659--8666, dec 2010.
\newblock ISSN 09574174.
\newblock \doi{10.1016/j.eswa.2010.06.065}.

\bibitem[Sweller et~al.(1998)Sweller, Merrienboer, and Paas]{22-Sweller1998}
Sweller, John, Merrienboer, Jeroen J G~Van, and Paas, Fred G W~C.
\newblock Cognitive architecture and instructional design.
\newblock \emph{Educational Psychology Review}, 10\penalty0 (3):\penalty0
  251--296, 1998.

\bibitem[Tallon-Baudry et~al.(1999)Tallon-Baudry, Kreiter, and
  Bertrand]{tallon1999sustained}
Tallon-Baudry, Catherine, Kreiter, Andreas, and Bertrand, Olivier.
\newblock Sustained and transient oscillatory responses in the gamma and beta
  bands in a visual short-term memory task in humans.
\newblock \emph{Visual neuroscience}, 16\penalty0 (03):\penalty0 449--459,
  1999.

\bibitem[Zeiler et~al.(2010)Zeiler, Krishnan, Taylor, and
  Fergus]{zeiler2010deconvolutional}
Zeiler, Matthew~D, Krishnan, Dilip, Taylor, Graham~W, and Fergus, Rob.
\newblock Deconvolutional networks.
\newblock In \emph{Computer Vision and Pattern Recognition (CVPR), 2010 IEEE
  Conference on}, pp.\  2528--2535. IEEE, 2010.

\bibitem[Zeiler et~al.(2011)Zeiler, Taylor, and Fergus]{Zeiler2011}
Zeiler, Matthew~D., Taylor, Graham~W., and Fergus, Rob.
\newblock {Adaptive deconvolutional networks for mid and high level feature
  learning}.
\newblock \emph{Proceedings of the IEEE International Conference on Computer
  Vision}, pp.\  2018--2025, 2011.
\newblock ISSN 1550-5499.
\newblock \doi{10.1109/ICCV.2011.6126474}.

\bibitem[Zeiler \& Fergus(2014)Zeiler and Fergus]{Zeiler2014}
Zeiler, Md and Fergus, Rob.
\newblock {Visualizing and understanding convolutional networks}.
\newblock \emph{Computer Vision (ECCV) 2014}, 8689:\penalty0 818--833, 2014.
\newblock ISSN 978-3-319-10589-5.
\newblock \doi{10.1007/978-3-319-10590-1{\_}53}.

\bibitem[Zhang \& LeCun(2015)Zhang and LeCun]{4-Zhang2015}
Zhang, Xiang and LeCun, Yann.
\newblock {Text Understanding from Scratch}.
\newblock \emph{arXiv preprint arXiv:1502.01710}, 2015.

\end{thebibliography}
\bibliographystyle{iclr2016_conference}

\newpage
\section*{Appendix}
\begin{figure}[tbh]
\begin{center}
\includegraphics[width=3.3in]{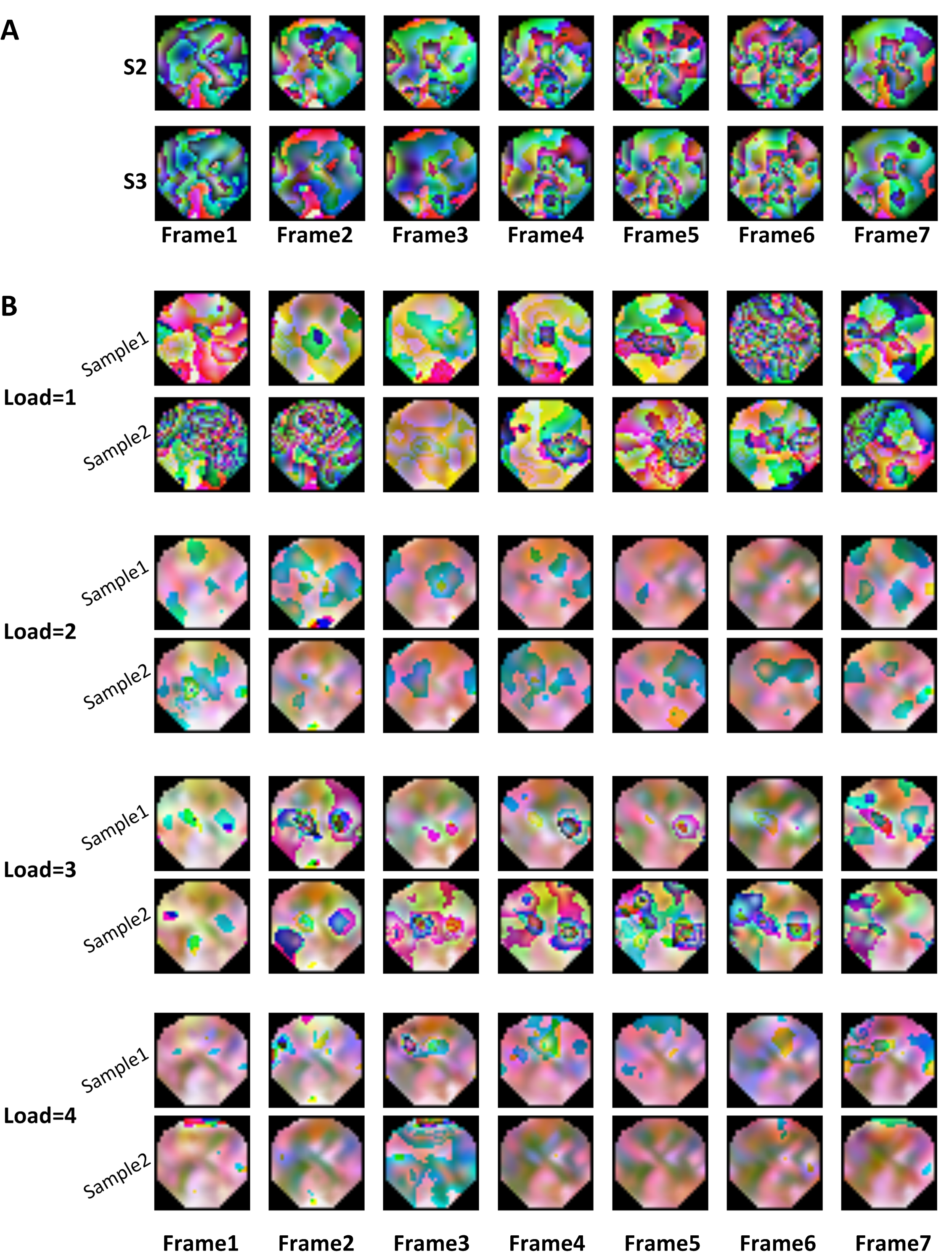}
\end{center}
\caption{A: Average frames obtained over multiple runs, under the same exact condition (same cognitive load level of the working memory task) from two different  subjects (S2 and S3).  B: multiple runs for the same condition (task and load level) for the same subject. For more detail, see section \ref{sec-exp}.
}
\label{figS1}
\end{figure}

An example demonstrating potentially high inter- and intra-subject variability of observed responses from different individuals performing same task, as well as different runs for the same subject performing the task several times, is shown in Figures \ref{figS1}a and \ref{figS1}b, respectively (for more details on this
experiment, see section \ref{sec-exp}).
More specifically, Figure \ref{figS1}a demonstrates the average frames obtained from two subjects within the same condition. Evidently, there are large inter-subject variations in the patterns emerging from average frames. Similarly, high variations could exist in responses recorded during multiple runs of the same task from the same subject, as shown in  Figure \ref{figS1}b.

\end{document}